\documentclass[12pt]{article}
\usepackage[utf8]{inputenc}

\usepackage{caption}
\usepackage{setspace}

\usepackage{hyperref}
\usepackage{subcaption}
\usepackage{numprint}
\usepackage{graphicx}
\usepackage{listings}
\usepackage{siunitx}

\usepackage[numbers]{natbib}
\usepackage{fancyhdr}
\usepackage{xcolor}

\usepackage{times}
\usepackage{microtype}
\usepackage{listings}
\usepackage{amsmath}
\usepackage{amsfonts}
\usepackage{amsthm}
\usepackage{amssymb}
\usepackage{numprint}
\usepackage{booktabs}
\usepackage{authblk}

\usepackage{placeins}

\newtheorem{definition}{Definition}
\usepackage{tikz}
\usetikzlibrary{shapes,arrows,shadows,decorations.markings}

\tikzstyle{bb}=[draw, fill=blue!20, text width=5em, 
    text centered, minimum height=3em,drop shadow]
\tikzstyle{pb}=[draw, fill=purple!20, text width=5em, 
    text centered, minimum height=3em,drop shadow]
\tikzstyle{rb}=[draw, fill=red!20, text width=5em, 
    text centered, minimum height=3em,drop shadow]
\tikzstyle{gb}=[draw, fill=olive!20, text width=5em, 
    text centered, minimum height=3em,drop shadow]
\tikzstyle{thinbox}=[draw, fill=teal!20, text width=4em, 
    text centered, minimum height=4.5em,drop shadow]
\tikzstyle{line} = [draw, -latex, length=3mm,width=5mm]

\newcommand{\beginsupplement}{%
        \setcounter{table}{0}
        \renewcommand{\thetable}{S\arabic{table}}%
        \setcounter{figure}{0}
        \renewcommand{\thefigure}{S\arabic{figure}}%
     }

\graphicspath{{./figures/}}
\newcommand{\btheta}{\boldsymbol\theta}
\newcommand{\x}{\mathbf{x}}
\newcommand{\X}{\mathbf{X}}
\newcommand{\Supp}{\text{Supp}}
\newcommand{\cond}{\mid}

\newcommand*\samethanks[1][\value{footnote}]{\footnotemark[#1]}

\title{Privacy-preserving data sharing via probabilistic modelling}
\date{}

\author{
  \textbf{Joonas Jälkö}$^1$\thanks{Correspondence to joonas.jalko@aalto.fi, Lead Contact}, \textbf{Eemil Lagerspetz}$^2$, \textbf{Jari Haukka}$^3$,\\
	\textbf{Sasu Tarkoma}$^2$,
	\textbf{Antti Honkela}$^{2}$\thanks{These authors contributed equally to this work as senior authors.}, 
	and \textbf{Samuel Kaski}$^{1,4}$\samethanks \\
	\vskip1ex
	$^1$ Helsinki Institute for Information Technology HIIT,\\
		Department of Computer Science, Aalto University, Finland \\
	$^2$ Helsinki Institute for Information Technology HIIT,\\
	Department of Computer Science, University of Helsinki, Finland\\
	$^3$ Department of Public Health, University of Helsinki, Finland\\
	$^4$ Department of Computer Science, University of Manchester, United Kingdom
}

\begin{document}
\maketitle

\section{Summary}

Differential privacy allows quantifying privacy loss resulting from accessing sensitive
personal data. Repeated accesses to underlying data incur increasing loss. Releasing data as
privacy-preserving synthetic data would avoid this limitation, but would leave open the problem of
designing what kind of synthetic data.
We propose formulating the problem of private data release through
probabilistic modelling. 
This approach transforms the problem of designing the synthetic data into choosing a model for the
data, allowing also including prior knowledge, which improves the quality of the synthetic data.
We demonstrate empirically, in an epidemiological study, that statistical discoveries can be
reliably reproduced from the synthetic data. We expect the method to have broad use in creating
high-quality anonymized data twins of key data sets for research.


\section{Introduction} 

Open release of data would be beneficial for research but is not feasible for sensitive data, for
instance clinical and genomic data. Since reliably anonymizing individual data entries is hard,
releasing synthetic microdata~\cite{Rubin1993} has been proposed as an alternative. To maximize
the utility of the data, the distribution of the released synthetic data should be as close as
possible to that of the original data set, but should not contain synthetic examples that are too
close to real individuals as their privacy could be compromised. Traditional methods of statistical
disclosure limitation cannot provide rigorous guarantees on the risk~\cite{Abowd2008}. However,
differential privacy (DP) provides a natural means of obtaining such guarantees.

DP \cite{dwork2006our, dwork_et_al_2006} provides a statistical definition of privacy
and anonymity. It gives strict controls on the risk that an individual can be identified
from the result of an algorithm operating on personal data. Formally, a randomized algorithm
$\mathcal{M}$ is $(\epsilon, \delta)$-DP, if for all data sets $X,X'$, where $X$ and $X'$ agree in
all but one entry, and for all possible outputs $S$ of $\mathcal{M}$, it satisfies
\begin{equation}
	\Pr(\mathcal{M}(X) \in S) \leq e^\epsilon \Pr(\mathcal{M}(X') \in S) + \delta,  
\end{equation}
\noindent
where $0\leq \delta <1$. The non-negative parameters $\epsilon, \delta$ define the strength of
the guarantee, with smaller values indicating stronger guarantees. Privacy is usually achieved by
introducing noise into the algorithms. DP has many desirable properties such as composability:
combining results of several DP algorithms is still DP, with privacy guarantees depending on how the
algorithms are applied \cite{dwork_et_al_2006, dwork2010boosting}. Another important property of DP
is invariance to post-processing \cite{DworkRoth}, which assures that the privacy guarantees of a
DP result remain valid after any post-processing. Thus we can use the results of a DP algorithm to
answer future queries and still have the same privacy guarantees.

Data sharing techniques under DP can be broadly separated into two categories as noted by Leoni
\cite{leoni2012non}: input perturbation, where noise is added to the original data to mask
individuals; and synthetic microdata, created from generative models learned under DP. The
input perturbation techniques lack generality as they are often suitable for only very specific
types of data, for example set-valued data \cite{chen2011publishing}. From now on we will focus
only on synthetic data based techniques. Using DP for releasing synthetic microdata provides a
more generalisable solution and was first suggested by Blum et.~al \cite{blum2008learning} for
binary data sets. Since then, multiple privacy-preserving data release techniques have been
proposed \cite{beimel2010bounds,chanyaswad2019ron, chen2012differentially, hardt2012simple,
Mohammed,xiao2010differentially, xiao2012dpcube}. However, the methods have so far been
limited to special cases such as discrete data \cite{beimel2010bounds, chen2012differentially,
gupta2012iterative, hardt2012simple, Mohammed,xiao2010differentially}, or having to draw a synthetic
data set from noisy histograms \cite{xiao2010differentially, xiao2012dpcube}. More recent work has
employed more powerful models~\cite{abay2018privacy,acs2018differentially, chanyaswad2019ron}.
These methods have been shown to be much more efficient and general compared to previous attempts.
However, these methods as well as other data sharing works share a limitation: they are not able to
use existing (prior) knowledge about the data set.

Typically the data sharing methods are build around a similar idea: learn a generative model from
the sensitive data under privacy guarantees and then sample a synthetic data set from the trained
model. These works differ mainly in the specific model used, and how the model is learned under DP.
Now one might ask is this not sufficient, if the model is a universal approximator (such as VAEs in
\cite{acs2018differentially}) and a sufficient amount of data is used to train it? The answer is
yes in principle, but in practice the amount of data required may be completely infeasible, as the
universal approximator would need to learn from data the structure of the problem, causality, and
all parameters. All this is made more difficult by capacity of the models being more limited
under DP, and the necessary tuning of hyperparameters coming with a privacy cost.

If the human modeller has knowledge of how the data have been generated, it is much more
data-efficient to put this knowledge to the model structure than to learn everything from scratch
with general-purpose data-driven models. For example, the data analyst might want to explicitly
model structural zeros, i.e. zeros that corresponds to an impossible outcome due to other features
of the data, e.g. alive subjects cannot have a cause of death. This is where the general purpose
models fall short. Instead of building a new general purpose model for private data sharing, we
propose a new essential component to private data sharing by augmenting the standard data sharing
workflow with a modelling task. In this modelling task, the user can encode existing knowledge of
the problem and the data into the model before the private learning, thus guiding the DP learning
task without actually accessing any private data yet.

We propose to give the modeller the tools of probabilistic modelling that provide a natural language
to describe existing knowledge about how the data have been generated. This includes any prior
knowledge which can be seamlessly integrated. In a continuous or high-dimensional data space there
is also another reason why probabilistic modelling is needed: finite data sets are often sparse and 
require smoothing that preserves the important properties of the data.



In this paper we formulate the principle of \emph{Bayesian DP data release}, which employs a
generative probabilistic model and hence turns synthetic data release into a modelling problem. We
demonstrate how the modelling helps in data sharing by using a general purpose model
as a starting point. We will increase the amount of prior knowledge encoded into the model and
show empirically how the synthetic data set becomes more similar to the original one when we
guide it with more prior knowledge. We show how the modelling becomes pivotal in making correct
statistical discoveries from the synthetic data. Code for applying the principle across model
families and data sets is available at
\url{https://github.com/DPBayes/twinify}\footnote{Code for experiments in the paper is available at
\url{https://github.com/DPBayes/data-sharing-examples}.}.


\section{Results}
\label{res_section}

\subsection{Overview of methods used in the experiments}

Our aim is to release a new synthetic data set that preserves the statistical properties of the
original data set while satisfying DP guarantees. Consider a data set $\X$ and a probabilistic model
$p(\X \cond \btheta)$ with parameters $\btheta$. We use the posterior predictive distribution (PPD)
$p(\tilde{\X} \cond \X)$,
\begin{equation}
	p(\tilde{\X} \cond \X) = \int_{\Supp(\btheta)} p(\tilde{\X} \cond \btheta) \, p(\btheta \cond \X) 
	\, \text{d} \btheta,
\end{equation}
\noindent
to generate the synthetic data. PPD tells us the probability of observing a new sample conditioned
on the data we have obtained thus far. Therefore, if our model sufficiently captures the generative
process, the PPD is the natural choice for generating the synthetic data. We sample the synthetic
data from the posterior predictive distribution, by first drawing $\tilde{\boldsymbol \theta}$ from
the posterior distribution $p(\btheta \cond \X)$ and then drawing new data point $\tilde{\x}$ from
the probabilistic model conditioned on $\tilde{\boldsymbol \theta}$, and repeating for all points.

\begin{figure}[!h]
    \centering
    \resizebox{0.8\textwidth}{!}{%
      \begin{tikzpicture}[node distance=0.22\textwidth, auto]
	\node (data) [rb] {Data ($\X$)};
	\node[thinbox, right of=data] (privacy) {Private learning};
	\node[bb, right of=privacy] (generator) {Generator $p_\X(\tilde{\X})$};
	\node[gb, right of=generator] (privdata) {Synthetic data ($\tilde{\X}$)};
	\draw[decoration={markings,mark=at position 1 with
    {\arrow[scale=3,>=stealth]{>}}},postaction={decorate}] (data)--(privacy);
	\draw[decoration={markings,mark=at position 1 with
    {\arrow[scale=3,>=stealth]{>}}},postaction={decorate}] (privacy)--(generator);
	\draw[decoration={markings,mark=at position 1 with
    {\arrow[scale=3,>=stealth]{>}}},postaction={decorate}] (generator)--(privdata);
\end{tikzpicture}
    }
    \caption{Standard differentially private data sharing workflow\label{fig:basic_pipeline}}
\end{figure}
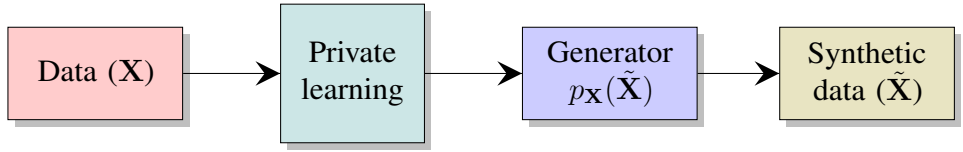

Many of the previous differentially private data sharing works share a common workflow, namely learn a
specific generative model from the data and share samples drawn from this generator. This pipeline is 
depicted in Figure \ref{fig:basic_pipeline}.

\begin{figure}[!h]
    \centering
    \resizebox{0.8\textwidth}{!}{%
      \begin{tikzpicture}[node distance=0.22\textwidth, auto]
	\node (data) [rb] {Data ($\X$)};
	\path (data.south)(0.0, -3.0) node (model) [pb] {Model(s) $p(\X,\btheta)$};
	\path (data.east)(0.22\textwidth, -1.5) node (privacy) [thinbox] {Private learning};
	\node[bb, right of=privacy] (generator) {Generator $p(\tilde{\X} \mid \X)$};
	\node[gb, right of=generator] (privdata) {Synthetic data ($\tilde{\X}$)};
	\draw[decoration={markings,mark=at position 1 with
    {\arrow[scale=3,>=stealth]{>}}},postaction={decorate}] (data)--(privacy);
	\draw[decoration={markings,mark=at position 1 with
    {\arrow[scale=3,>=stealth]{>}}},postaction={decorate}] (model)--(privacy);
	\draw[decoration={markings,mark=at position 1 with
    {\arrow[scale=3,>=stealth]{>}}},postaction={decorate}] (privacy)--(generator);
	\draw[decoration={markings,mark=at position 1 with
    {\arrow[scale=3,>=stealth]{>}}},postaction={decorate}] (generator)--(privdata);
\end{tikzpicture}
    }
    \caption{Bayesian DP data release\label{fig:new_pipeline}}
\end{figure}
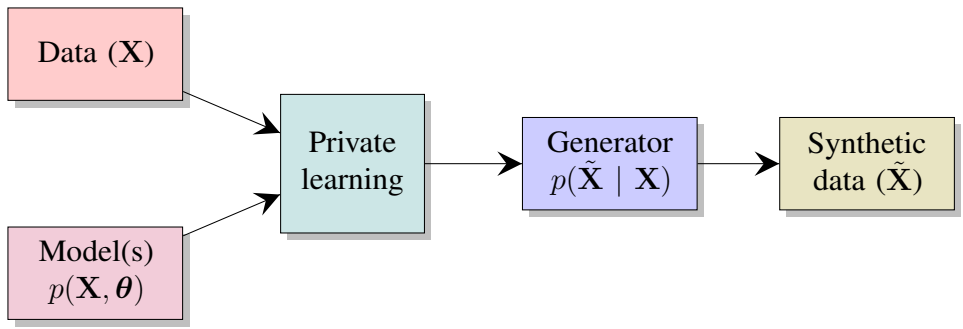

What we suggest is to augment this pipeline with domain knowledge of the data holder. This is 
possible through probabilistic modelling which gives a natural language for encoding such knowledge
prior to learning. In out experiments, we have used the new improved  pipeline, depicted in 
Figure \ref{fig:new_pipeline}.

\subsection{Reproducing statistical discoveries from the synthetic data}

In order for private data sharing to be useful, we need to retain important statistical information
in the synthetic data while preventing re-identification of data subjects. Next we will demonstrate
how encoding prior knowledge becomes essential in making correct statistical discoveries from the
synthetic data.

To test whether the same discoveries can be reproduced from the synthetic as from the
original data set, we generated a synthetic replica of a data set used in an epidemiological
study~\cite{but2017cancer}, using a general-purpose generative model family (mixture model). Prior
to learning, we encoded experts' domain knowledge about the data into the probabilistic model.

The data have previously been used to study the association between diabetes and alcohol
related deaths (ARD) using a Poisson regression model~\cite{niskanen2018excess}. The study
showed that males and females exhibit different behaviour in terms of alcohol related
mortalities. We encoded this prior knowledge into the model by learning
independent mixture models for males and females. Another type of prior knowledge we had, comes from
the nature of the study that produced the data: the data of each subject ends either on a specific
date, or at death. Hence, the status at the endpoint is known to have a one-to-one correspondence
on certain features such as duration of the follow-up and most importantly the binary indicator
that tells if individual died of alcohol related causes. We encoded this prior knowledge into the
probabilistic model as well. For details on the models we refer to the Materials and Methods.

After building the model, we learned the generative model under DP and generated the
synthetic data. We fit the same Poisson regression model that was used in the earlier study
\cite{niskanen2018excess} to the synthetic data as well, and compared the regression coefficients of
the two models.

\begin{figure*}[!h]
	\begin{subfigure}[t]{0.49\textwidth}
		\includegraphics[width=\linewidth]{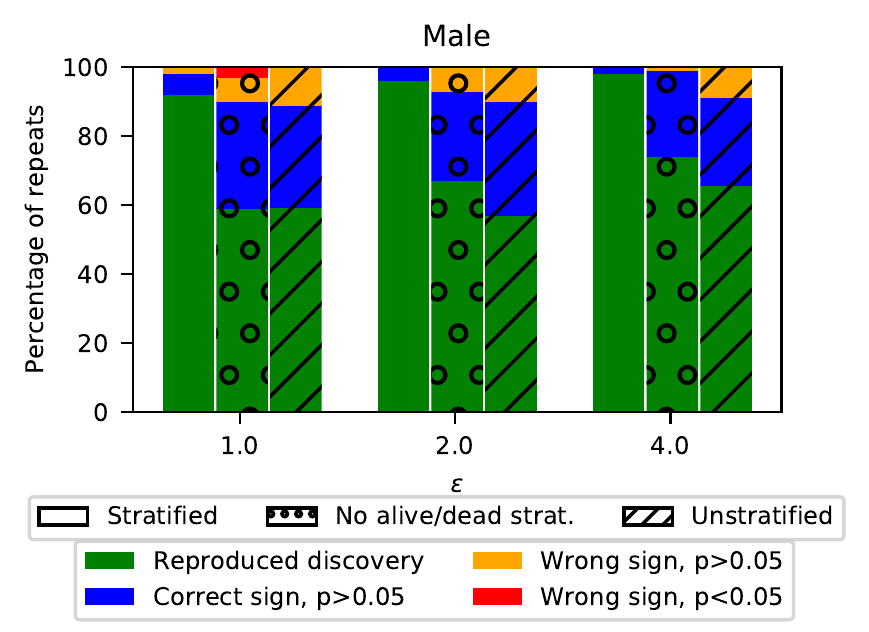}
		\caption{}
	\end{subfigure}
	\begin{subfigure}[t]{0.49\textwidth}
		\includegraphics[width=\linewidth]{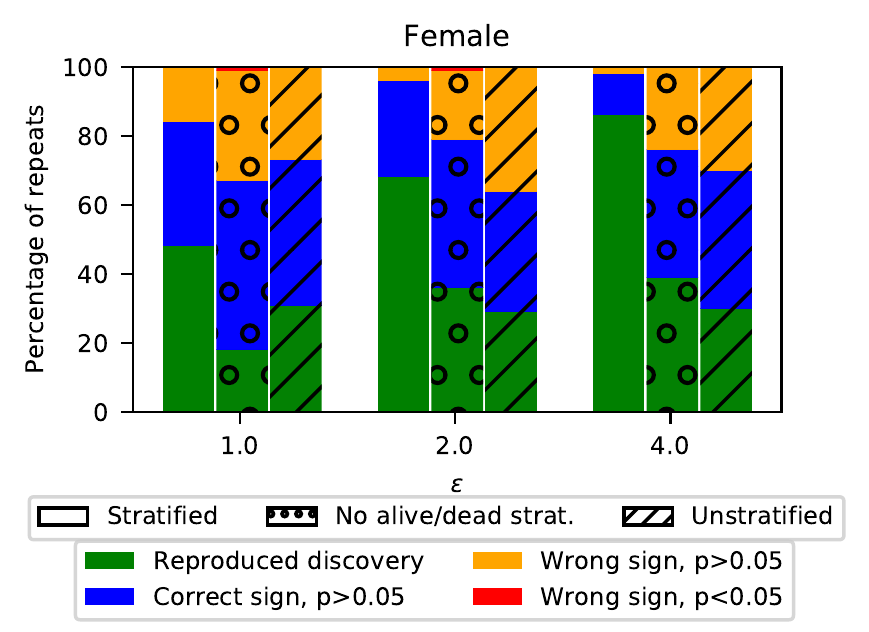}
		\caption{}
	\end{subfigure}
	\caption{\textbf{ARD study: encoding prior knowledge into the generative model improves performance}. 
			For both males (a) and females (b) we recover the correct statistical discovery with high
			probability when we guide the model sufficiently with prior knowledge. The prior knowledge is
			increased from right to left in both a and b. In ''Stratified'', we have independent mixture models
			for the genders and deterministic features due to studys outcome. In ''No alive/dead strat.'' we
			have independent models for the genders, and in ''Unstratified'' we treat all features within a
			mixture component as independent. For a reproduced discovery, we required the association between
			ARD and medication type to be found for all medication types with significance ($p<0.05$). The
			figures show results of 100 independent repeats of each method with three levels of privacy
			(parametrized by $\epsilon$). \label{fig:both_strat_bars}
	} 
\end{figure*}

From the synthetic data, we make two key observations. \textbf{(1)} We can reproduce the discovery
that the diabetics have a higher risk for ARD than the non-diabetics, which agrees with the
previous results on the original data \cite{niskanen2018excess}. The bar dubbed ''Stratified'' in
Figure \ref{fig:both_strat_bars} shows that we can reproduce the discoveries with high probability
for males with relatively strict privacy guarantees ($\epsilon=1$). For females we need to
loosen the privacy guarantees to $\epsilon=4$ in order to reproduce the statistical discovery
with high probability. We discuss the difference between males and females in the next Section.
\textbf{(2)} In order to reproduce the discovery, we need to have the correct model. Figure
\ref{fig:both_strat_bars} shows results of three different models: ''Stratified'' equipped with
prior knowledge on gender and outcome of the follow-up, ''No alive/dead strat'' with prior knowledge
only on gender and ''Unstratified'' without either type of prior knowledge. We see that the more
prior knowledge we encode into the model, the better reproducibility we get. For males, with
strict privacy ($\epsilon=1$) we increase the rate of reproducibility almost by $40\%$ by having
the correct model. For females, the effect is even stronger, however best visible with larger
$\epsilon$.

\subsection{Performance of DP data sharing}

Next we will demonstrate the usability as well as the limitations of the proposed general DP data 
sharing solution.


\paragraph{DP data sharing works best when data are plentiful.}

\begin{figure*}[!h]
	\centering
	\includegraphics{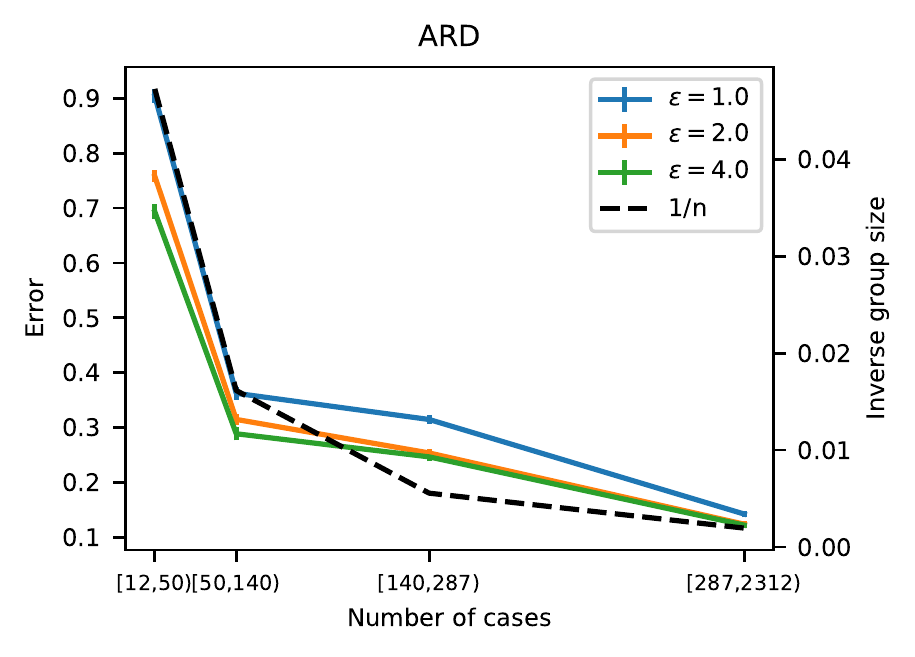}
	\caption{
		\textbf{Accuracy of findings from synthetic data as a function of their rarity, ARD study}.
			Accuracy of regression coefficients learned from synthetic data rapidly improves as the number
			of relevant samples grows. The solid curves show mean absolute error within a prevalence bin
			between the regression coefficients learned from original and synthetic data. Figure shows average
			result over 100 independent runs of the algorithm. The dashed line is proportional to the expected
			behaviour of an optimal estimator (see text); note the different scale in the y-axis (shown on the
			right) Results are from the stratified model. Tickmarks on the x-axis are (min, max) number of
			relevant samples within the respective bin. Error bars denote the standard error of mean. Results
			shown for three values of the privacy parameter $\epsilon$. \label{fig:rarity_vs_acc}
		} 
\end{figure*}

As we saw in Figure \ref{fig:both_strat_bars}, the utility is better for males than the females,
especially for strict privacy guarantees. To understand the difference between the two cases (males,
females) in the ARD study, we note the much smaller sample size for ARD incidences among females
(\numprint{520} vs \numprint{2312}). Since DP guarantees indistinguishability among individuals in
the data set, it is plausible that the rarer a characteristic, the less well it can be preserved in
DP-protected data. To assess whether this holds for the regression coefficients in the ARD study,
we divided the regression coefficients, both male and female, into four equal-sized bins based on
how many cases exhibited the corresponding feature and computed the mean absolute error between the
original and synthetic coefficients within these bins. Figure \ref{fig:rarity_vs_acc} shows that
the regression coefficients with higher number of cases are more accurately discovered from the
synthetic data.

Previously Heikkilä \emph{et al.}~\cite{heikkila17} showed that the error of estimating parameter
mean under $(\epsilon, \delta)$-DP decreases proportional to $\mathcal{O}(1/n)$, where $n$ is the
size of the data set. Figure \ref{fig:rarity_vs_acc} shows that the error in the ARD study follows
closely the expected behaviour as the number of cases increases. In this experient, the inverse
group size was estimated with the average of the inverse group sizes within a bin.

However, the data size is not the only determining factor for the utility of DP data sharing. Next
we will show how more clear-cut characteristics of the data are easier to discover, even with fewer
samples.

\paragraph{Picking up a weak statistical signal is difficult for DP data sharing.}

\begin{figure*}[h!]
	\centering
	\begin{subfigure}[t]{0.49\textwidth}
		\includegraphics{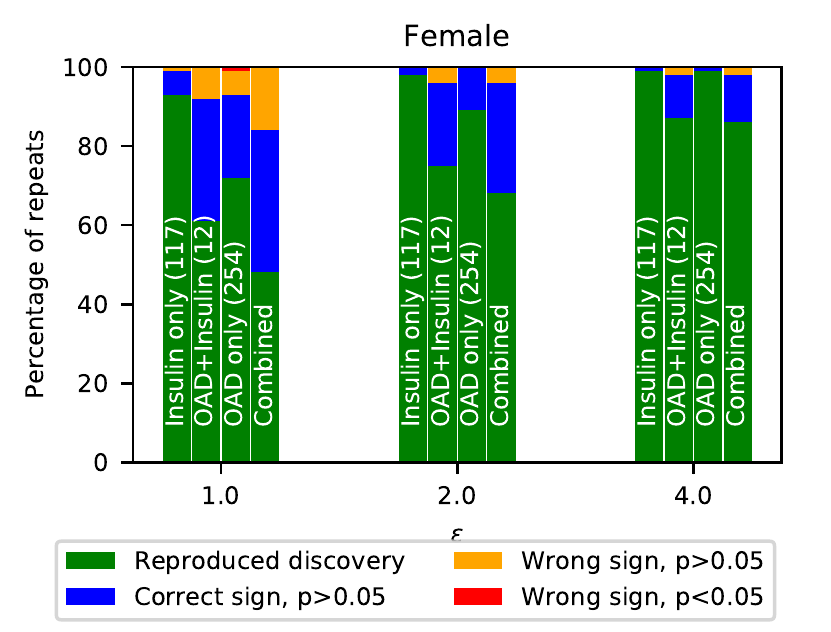}
		\caption{\label{fig:conclusions_female}} 
	\end{subfigure}
	\hfill
	\begin{subfigure}[t]{0.49\textwidth}
		\includegraphics{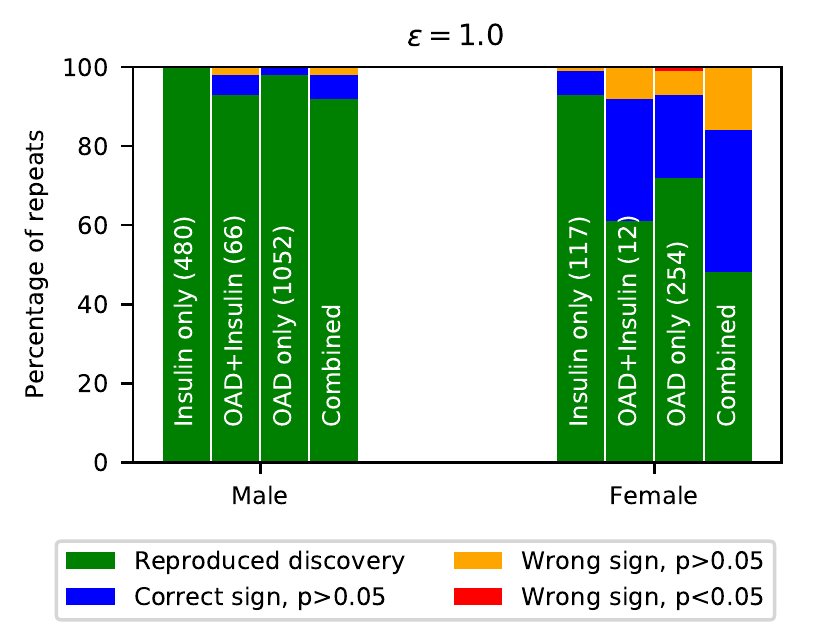}
		\caption{\label{fig:conclusions}}
	\end{subfigure}
	\caption{
		The statistical signal is weaker in female data (ARD study).
		\textbf{(a) Likelihood of reproducing findings as a function of privacy guarantee. 
		Female case}.
		The statistical discoveries are reliably reproduced from the synthetic data for the strictest
		privacy requirements. Results for combined case and for each subgroup separately. Combined results:
		all subgroups are required to have the correct sign and $p < 0.05$ to call the discovery reproduced.
		The size of each subgroup is shown in parenthesis. Results are from stratified model.
		\textbf{(b) Likelihood of reproducing findings from synthetic data}. 
		For males (\numprint{226372} samples), the discoveries can be reproduced with high probability
		from the synthetic data. For females (\numprint{208148} samples), the probability of reproducing
		discoveries is lower. Bars show discoveries for each type of diabetes medication separately, and for
		all combined. In the combined case, for a reproduced discovery, we required the association between
		ARD and medication type to be found for all medication types with significance ($p<0.05$). Results
		of 100 independent repeats of the method with privacy level $(\epsilon=1.0, \delta=10^{-6})$ using
		the stratified model.
		}
\end{figure*}

The ARD study stratifies individuals based on three types of diabetes treatment: Insulin only,
orally administered drug (OAD) only and Insulin+OAD treatment. Each of these therapies is
treated as an independent regressor. For a reproduced discovery, we require that all of the
regressors are positive and have sufficient statistical significance ($p<0.05$). From Figure
\ref{fig:conclusions_female} we see that the probability of reproducing the discoveries for
each subgroup increases as $\epsilon$ grows. However, we also see that for the ''Insulin only''
subgroup we recover the correct discovery with higher rate compared to the larger subgroup ''OAD
only''. The reason, why the smaller subgroup ``\emph{Insulin only}'' is more often captured with
sufficient significance than the largest subgroup ``\emph{OAD only}'', can be explained by the
original regression coefficients shown in Table \ref{table:coef_table}. The OAD only subgroup has a
significantly smaller effect on the ARD than the Insulin only, thus making it more difficult for the
mixture model to capture it. However as we increase $\epsilon$, the correlation between OAD only and
ARD is more often captured. Both of these effects are also visible in the male case, as we see from
Figure \ref{fig:conclusions}, however in a smaller scale.

Some of the regression coefficients learned from the synthetic data diverge from the ground
truth, which seems to also persists without privacy (see Column $\epsilon=\infty$ in Table
\ref{table:coef_table}). In our experiments we have used a small number of mixture components
($k=10$) as a compromise between sufficiently high resolution (we can make correct statistical
discoveries) and private learning that becomes more difficult as the number of parameters grow.
Increasing the number of mixture components resolves this inconsistency by improving the fit in
non-private case (See Table S1 in Supplemental Information)

\begin{figure*}[!t]
	\centering
	\includegraphics{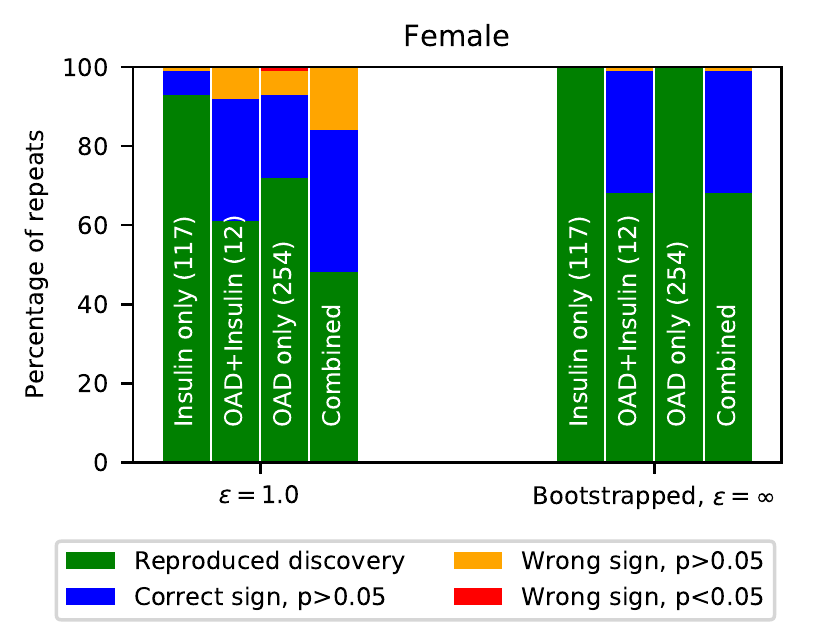}	
	\caption{
	\textbf{ARD study:} 
	The statistical signal is weak in female data, and discoveries cannot be made 
	with sufficient significance. On the left, the bars show results for private synthetic data with 
	$\epsilon=1.0$ of 100 independent runs using the stratified model. On the right, the bars show 
	results for $100$ times bootstrapped original data.
	\label{fig:female_bootstrap}}
\end{figure*}

To evaluate the strength of the statistical signals in the female ARD study, we ran the Poisson
regression study with bootstrapped original female data. Figure \ref{fig:female_bootstrap} shows
that under $100$ bootstrap iterations, $\approx 30 \%$ of the repeats did not reach the required
statistical signicance. This shows that the statistical signal in female data is weak to begin with,
and therefore may be difficult for data sharing model to capture.

\begin{table*}[t]
	\centering
	\textbf{Female}
	\resizebox{\textwidth}{!}{%
	\begin{tabular}{llllllll}
\toprule
{} &   Coefficient & Number of cases & Original coef. $\pm$ Std. Error &     $\epsilon=1.0$ &     $\epsilon=2.0$ &     $\epsilon=4.0$ &  $\epsilon=\infty$ \\
\midrule
0 &      OAD only &             254 &               $0.657 \pm 0.108$ &  $0.303 \pm 0.197$ &  $0.474 \pm 0.209$ &  $0.591 \pm 0.189$ &  $0.887 \pm 0.149$ \\
1 &   OAD+Insulin &              12 &               $0.873 \pm 0.304$ &  $0.658 \pm 0.516$ &   $0.846 \pm 0.44$ &  $1.074 \pm 0.427$ &  $1.124 \pm 0.366$ \\
2 &  Insulin only &             117 &                $1.68 \pm 0.135$ &   $0.91 \pm 0.379$ &  $1.085 \pm 0.312$ &  $1.313 \pm 0.293$ &  $1.521 \pm 0.206$ \\
\bottomrule
\end{tabular}

	}
	\vskip6pt
	\textbf{Male}
	\resizebox{\textwidth}{!}{%
	\begin{tabular}{llllllll}
\toprule
{} &   Coefficient & Number of cases & Original coef. $\pm$ Std. Error &     $\epsilon=1.0$ &     $\epsilon=2.0$ &     $\epsilon=4.0$ &  $\epsilon=\infty$ \\
\midrule
0 &      OAD only &            1052 &               $0.435 \pm 0.049$ &  $0.412 \pm 0.166$ &  $0.502 \pm 0.152$ &   $0.538 \pm 0.12$ &  $0.532 \pm 0.089$ \\
1 &   OAD+Insulin &              66 &               $0.582 \pm 0.129$ &  $0.748 \pm 0.304$ &  $0.816 \pm 0.282$ &  $0.858 \pm 0.234$ &   $0.864 \pm 0.17$ \\
2 &  Insulin only &             480 &               $1.209 \pm 0.063$ &  $1.033 \pm 0.189$ &  $1.188 \pm 0.205$ &  $1.257 \pm 0.138$ &  $1.262 \pm 0.123$ \\
\bottomrule
\end{tabular}

	}
	\caption{\textbf{ARD study}, \textbf{ABOVE} : Females, \textbf{BELOW} : Males.
	The magnitude of the statistical effect in the male case is well preserved in synthetic data.
	DP and synthetic non-DP ($\epsilon = \infty$) results are average over 100 runs, error denoting the 
	standard deviation. The error in original coefficients shows the standard error for the regression model.
	}
	 \label{table:coef_table}
\end{table*}

Despite DP data sharing having difficulties with weak statistical signal and limited data, it
provides an efficient solution for privacy-preserving learning, especially when we are not certain
about the future use of the data. Next we will discuss how DP-based synthetic data stands against
traditional query based DP approaches.


\paragraph{Performance against tailored mechanism.}
As discussed, one of the greatest advantages of releasing a synthetic data set is that it can be
used in arbitrary tasks without further privacy concerns. Using traditional DP techniques, a data
holder that wants to allow DP access to a sensitive data set needs to set a privacy
budget at the desired level of privacy, and split this budget for each access that the data
is subjected to. As soon as the privacy budget runs out, the data cannot be used in any additional
analysis without unacceptable privacy risk.

We will next show that the data sharing methods can outperform traditional DP techniques, if the
data are to be accessed multiple times. We evaluate the performance on two data sets, a mobile phone
app data set \cite{oliner2013carat} referred to as Carat, and the publicly available set of US
Census data ''UCI Adult'' \cite{Dua:2019}. As data sharing methods we apply a mixture model based
PPD sampling method (''Mixture model'') and a Bayes networks based method PrivBayes~\cite{Zhang}
(''Bayes network'').

\begin{figure*}[!t]
	\centering
	\begin{subfigure}[t]{0.49\textwidth}
		\includegraphics{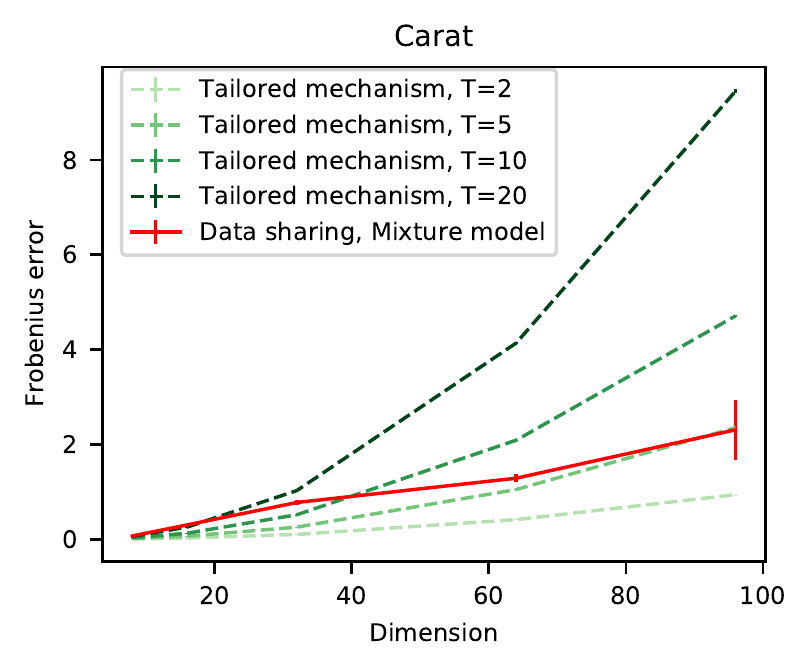}
		\caption{
		\label{fig:tailored_vs_sharing}}
	\end{subfigure}
	\hfill
	\begin{subfigure}[t]{0.49\textwidth}
		\includegraphics{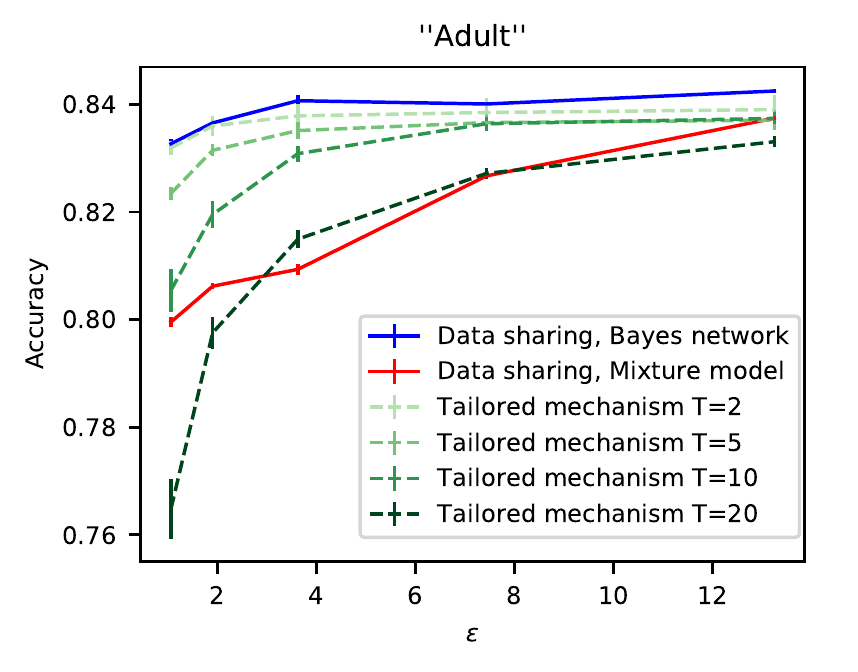}
		\caption{
		\label{fig:adult_pb_vs_sharing}}
	\end{subfigure}
	\caption{Performance against tailored mechanisms
	\textbf{(a) Carat study}.
		The data sharing method outperforms the tailored mechanism as the number of anticipated future
		queries (T) grows, in terms of classification accuracy. Curves show the Frobenius norm between
		original and synthetic covariance matrices. Privacy budget was fixed to $(1.0, 10^{-5})$. Average of
		10 runs. Errorbars denote the standard error of mean.
	\textbf{(b) Adult study}.
		Synthetic data from the Bayes network model outperforms the tailored mechanism. While a tailored
		mechanism is more accurate for loose privacy guarantees (large $\epsilon$) and few queries (small
		T), also the mixture model based data release is more accurate for multiple queries and tighter
		privacy guarantees. Average classification accuracy over 10 independent runs. Error bars denote
		standard error of mean.
	}
\end{figure*}

\begin{figure*}[!t]
	\centering
	\begin{subfigure}{0.4\textwidth}
      \includegraphics{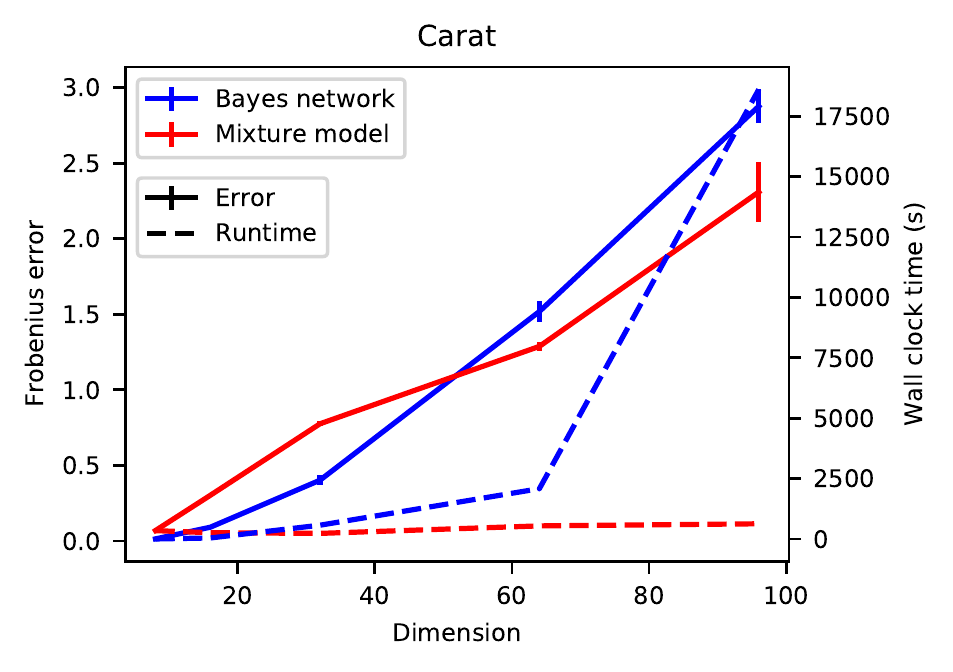}	
	  \caption{
	  \label{fig:dpvi_vs_pb_carat}
	  }
	\end{subfigure}
	\hfill
	\begin{subfigure}{0.4\textwidth}
	  \includegraphics{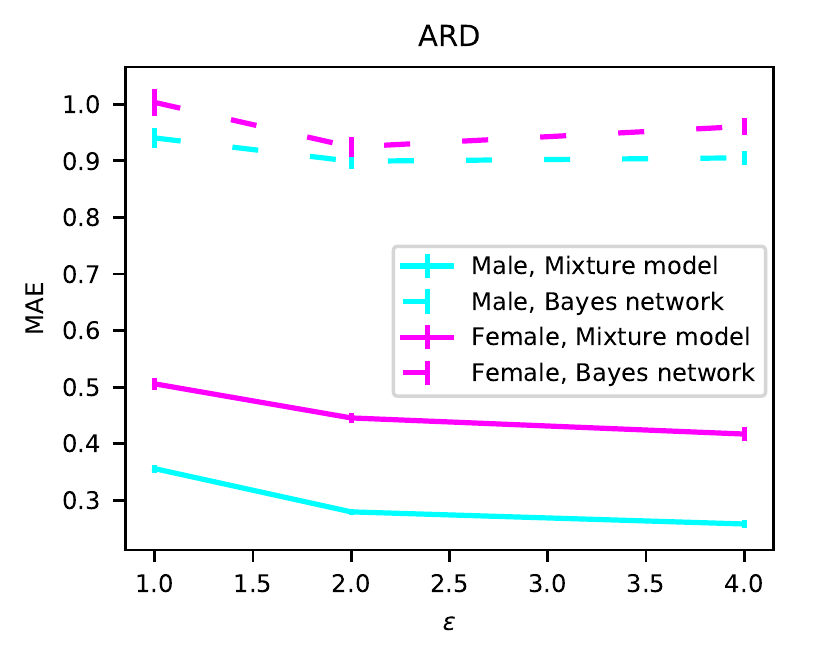}
  	  \caption{	
	  \label{fig:diabetes_vs_pb}
	  }	
	\end{subfigure}
	\caption{Comparing Mixture model and Bayes networks using two different data.
	  \textbf{(a) Accuracy and computation speed of two models in generating synthetic data (Carat study)}.
	  For low-dimensional discrete data Bayes networks are good, but as
	  dimensionality grows their computation time becomes intolerable and mixture
	  models more accurate. The solid lines denote the mean Frobenius norm
	  \eqref{eq:frob_norm} between the original and synthetic covariance matrices, with
	  error bars denoting standard error of the mean from 10 independent runs of the
	  algorithm. The dashed lines show the runtimes. Privacy budget was fixed to
	  $(\epsilon=1.0, \delta=10^{-5})$.
	  \textbf{(b) Accuracy of data synthesized with two models (ARD study)}.
	  Mixture models preserve regression coefficients better than the Bayes network.
	  The curves show mean absolute error between the original and the learned coefficients.
	  Average over 100 runs. Error bars: standard error of the mean.
	  }
\end{figure*}

Consider that the data holder splits the budget uniformly among $T$ anticipated queries. Figure
\ref{fig:tailored_vs_sharing} illustrates how the number of anticipated queries will affect the
accuracy. We compared the data sharing method against perturbing the covariance matrix with Gaussian
noise, according to the Gaussian mechanism \cite{dwork2006our} (``tailored mechanism''). We measured
the accuracy in terms of the Frobenius norm (see Equation \ref{eq:frob_norm}) between the true and
the DP covariance matrices. Already with $T=10$ queries, releasing a synthetic data set outperforms
the tailored mechanism for this high-dimensional data. We show results only for the mixture model 
because the difference in performance between the mixture model and the Bayes networks is small in 
this example (See Figure
\ref{fig:dpvi_vs_pb_carat})

As another example, we compared the synthetic data release on the Adult data against private
logistic regression classifier~\cite{jalko2017differentially}. Figure \ref{fig:adult_pb_vs_sharing}
shows that the Bayes network consistently outperforms the tailored mechanism, and for strict privacy
requirement (small $\epsilon$) also the mixture model performs better than the tailored mechanism
given 20 or more queries.


\paragraph{Demonstration on two parametric families of distributions}
Finally, we will demonstrate the results from two data sharing approaches using two very different
universal probilistic models making different computational tradeoffs. We evaluate the performance
between mixture models and Bayes networks on the ARD, Carat and Adult data sets.

For the Carat data, Figure \ref{fig:dpvi_vs_pb_carat} shows that the Bayes network is accurate when
the dimensionality of the data is low, but as the dimensionality grows, synthetic data generated
from the mixture model achieves higher accuracy than data from Bayes networks, which also becomes
computationally exhausting as the dimension increases. From Figure \ref{fig:dpvi_vs_pb_carat}, we
can see that learning the mixture model takes only a fraction of the Bayes networks computational
time. Similarly, in the ARD study, the mixture models outperforms Bayes networks (Figure
\ref{fig:diabetes_vs_pb}).

As a final comparison between the Bayes networks and mixture model, we compared the two in the
previously introduced classification task using the Adult data set, which has fewer samples
compared to ARD and Carat data (Adult \numprint{30162} samples, Carat \numprint{66754} samples and
ARD Females \numprint{208 148}, ARD Males \numprint{226 372}). After learning the
generative model, we used the synthetic data obtained from the generative model to train a logistic
regression classifier and demonstrated the performance by predicting income classes. Figure
\ref{fig:adult_pb_vs_sharing} illustrates that in this example, the Bayes networks outperforms the
mixture model in terms of classification accuracy.


\section{Discussion}
Dwork et al.~\cite{dwork2009complexity} showed theoretically that there is no computationally
efficient DP method for data sharing that would preserve all properties of the data. They consider
the problem from the learning theory perspective, where the aim is to accurately answer a set of
queries. Accurate answers become infeasible as the size of this query set grows. However, if we only
need to preserve the most important properties of the data, the set of queries we want to accurately
answer stays bounded in size, giving a way out. We argue that it would already be highly useful to be
able to answer questions of the important properties; and moreover, the bigger picture may be more
relevant than all the unique characteristics in the data.

As we saw in the Adult example, the DP data release can perform as well as the tailored mechanism
even when answering just one query, and progressively better for multiple queries. However, as our
experiments exemplify, encoding of the prior knowledge has a significant impact on the results.
In fact, what we are proposing is to transform the DP data release problem into a modelling problem,
which includes as an essential part the selection of the model according to the data and task, and 
bringing in available prior knowledge.

We illustrated in Figure \ref{fig:rarity_vs_acc} how increasing the number of relevant samples
improves the results. As is common with all differentially private methods, the data release works
better when the original data set has a large number of samples. This is because of the nature of DP;
it is easier to mask the contribution of one element of the data set when the number of samples is
large.

Recently, Karwa et al.~\cite{karwa2018finite} showed that DP has a broadening effect on the
confidence intervals of statistical quantities learned under DP. Their proof was for Gaussian mean
estimation; however, intuitively this property should translate to other differentially private
tasks as well. The width of the confidence intervals depends on both the required level of privacy
and the number of samples. This suggests that we should not expect to
necessarily reproduce all the same discoveries under DP.

In the past, there has been discussion on whether standard random number generators (RNG) can be
used to assure DP \cite{garfinkel}. In the actual data release setting we would need to consider
using cryptographically secure RNGs to properly provide individuals in the data set the DP
guarantees. Also, limited accuracy of floating point arithmetics make it possible for an attacker to
break DP due to errors in approximation~\cite{Mironov2012}. However, these problems are by no means
specific to DP data release but apply to all DP methods.

One major question for all DP algorithms is how to set the privacy
parameters $\epsilon$ and $\delta$. While the parameters are in
principle well-defined, their interpretation depends for example on
the chosen neighbourhood relation. Furthermore, the parameters are
worst-case bounds that do not fully capture for example the fact that
we do not release the full generative model but only samples drawn
from the posterior predictive distribution. Our use of
$\epsilon \approx 1$ is in line with widely accepted standards derived
from observed feasibility of membership inference attacks. Given the
complicated relationship between the released and original data, it
seems unlikely that the privacy of specific data subjects could be
compromised in this setting under this privacy level.

\section{Conclusions}

We have reformulated the standard differentially private data sharing by formulating it
as a modelling task. Using probabilistic modelling, we can express prior knowledge about the data
and processes that generated the data before the training, thus guiding the model towards right
directions without additional privacy cost. This makes it possible to extend the DP data sharing
solution to data sets that are of limited size, but for which there exists domain knowledge.

Differentially private data sharing shows great potential, and would be particularly useful for
data sets which will be used in multiple analyses. Census data is a great example of such data.
Also, as private data sharing allows arbitrary downstream tasks with no further privacy cost,
it is a good alternative for tasks for which there is no existing privacy-preserving counterpart. 

Our results demonstrate the importance of guiding the data sharing task with prior knowledge about the 
data domain, and that when this prior knowledge is encoded into the probabilistic model, the synthetic 
data maintains the usability of the original data in non-trivial tasks.


\section{Materials and methods}
\subsection{Materials}

For the ARD study \cite{but2017cancer}, the data came from \numprint{208148} females and
\numprint{226372} males and comprised of three continuous, five binary and two categorical features.

Carat data set: Carat~\cite{oliner2013carat} is a research project that maintains a mobile phone
app that helps users understand their battery usage. We obtained a subset of Carat data from the
research project. Our aim was to privately release a data set that consists of installed apps of
\numprint{66754} Carat users. In order to have some variance in the data, we dropped out the 100
most popular apps that were installed on almost every device and used the 96 next most popular apps
to subsample in the experiments.

In the Adult study of the UCI machine learning repository \cite{Dua:2019}, we trained the generative
model with \numprint{30162} samples with 13 features of both continuous and discrete types. Separate
test set consisted of \numprint{15060} instances, out of which $75.4\%$ were labelled $<$50k\$.

\subsection{Differential privacy}

In our experiments, we have used approximate differential privacy, definition given below.

\begin{definition}[Approximate differential privacy \cite{dwork2006our}]
	\label{dp_definition}
	A randomized algorithm $\mathcal{M} : \mathcal{X}^N \rightarrow \mathcal I$ satisfies 
	$(\epsilon, \delta)$ differential privacy, if for all adjacent datasets 
	$X, X' \in \mathcal{X}^N$ and for all measurable $S \subset \mathcal{I}$ it holds that
	\begin{align}
		\Pr(\mathcal{M}(X) \in S) \leq e^\epsilon\Pr(\mathcal{M}(X') \in S) + \delta.
	\end{align}
\end{definition}
\noindent
We consider data sets as adjacent in the substitute relation, i.e. if we get one by replacing a single
element of the other and vice versa. The privacy parameter $\delta$ used in the experiments was set to
$10^-6$ for the ARD study and $10^-5$ for both Carat and Adult studies.


\subsection{Probabilistic models}

\paragraph{Mixture model} \label{mm_intro} 

Mixture model is an universal approximator of densities. The probability density for mixture model
with $K$ mixture components is given as 
\begin{equation}
	p(\X \cond \btheta, \boldsymbol{\pi}) = \sum_{k=1}^K \boldsymbol{\pi}_k p(\X \cond \btheta^{(k)}).
\end{equation}
It allows capturing complex dependency structures through the differences between less complex
mixture components (the densities $p(\X \cond \btheta^{(k)})$). There is no limitation on what
kinds of distributions can be used for the mixture components, and thus a mixture model is suitable
for arbitrary types of data. In this work we assume independence of features within each mixture
component. This means that the component distribution factorizes over the features, and we can write
\begin{equation}
	\label{eq:ind_mm}
	p(\X \cond \btheta, \boldsymbol{\pi}) = \sum_{k=1}^K \boldsymbol{\pi}_k \prod_{j=1}^D p(\X_j \cond \btheta^{(k)}_j),
\end{equation}
where the $\X_j, j = 1,\ldots,D$ denote the $D$ features of the data and $\btheta^{(k)}_j$ the parameters
associated with the $j$th feature of the $k$th component distribution.
Intuitively the problem can be seen as finding clusters of features such that each cluster has a
axis-aligned covariance structure. As the number of such clusters increases, we can cover the data more
accurately.

In our experiments with mixture models, we used posterior predictive distribution as the generative
model. The only access to data is through the posteriors of the model parameters, which we
learned under DP using the DPVI method \cite{jalko2017differentially}. DPVI learns a mean field
approximation for the posterior distributions of model parameters using DP-SGD \cite{Abadi2016}.
The number of mixture components $K$ was set to 10 for data with fewer dimensions ($<20$) and to 20
for data with more dimensions ($\geq 20$). If necessary, this number, along with hyperparameters of
DPVI, could be optimized under DP \cite{Liu2019}, with potentially significant extra computational
cost.

\paragraph{Bayes networks} 

A Bayes network is a graphical model that presents the dependencies across random variables as a
directed acyclic graph (DAG). In the graph, the nodes represent random variables and the edges
dependencies between the variables. To learn the graphs privately and to sample the synthetic data,
we used the PrivBayes method \cite{Zhang} which builds the graph between the features of the data,
and no additional latent variables were assumed. The topology of the network is chosen under DP by
using the exponential mechanism \cite{mcsherry2007mechanism}, and the conditional distributions that
describe the probability mass function are released using Laplace mechanism \cite{dwork_et_al_2006}.


\subsection{Model details}
For the mixture model, we need to choose how to model each feature in the data sets. In all our
experiments we used the following distributions: Continuous features were scaled to the unit
interval and modelled as Beta distributed. The parameters for Beta-distributed variables were given
a Gamma$(1,1)$ prior. Discrete features were modelled as either Bernoulli or Categorical random
variables based on the domain. Both in Bernoulli and Categorical cases, the parameters were given a
uniform prior. Table \ref{mm_details} summarizes the mixture models used in the experiments.

\begin{table}[!h]
	\centering
    \begin{tabular}{ | l | l | l | p{4cm} |}
    \hline
    Dataset & $K$ & Variable types & Details \\ \hline
		ARD & 10 & Binary, Categorical, Beta &  Separate mixture models for males and females and also separation
		based on outcome of the follow-up. \\ \hline
	Carat & 20 & Binary &  Within a mixture component, the features were treated as independent.\\ \hline
	Adult & 10 & Binary, Categorical, Beta &  Separate mixture models for high/low income. ''Hours per week'', ''Capital Loss'' and ''Capital Gain'' features discretized into 16 bins.\\
    \hline
    \end{tabular}
	\caption{Summary of mixture model details. \label{mm_details}}
\end{table}


\subsubsection{Prior knowledge used in the ARD study}
\label{sec:ard_details}
In the ARD study, we showed how incorporating prior knowledge into the model improves the utility
of data sharing. Next we will describe in detail the type of knowledge we used to model the data.
We will encode the prior knowledge into the mixture model given in Equation \eqref{eq:ind_mm}.
This corresponds to the model referred to as ''Unstratified'' in Figure \ref{fig:both_strat_bars}.

We start by splitting the probabilistic model based on gender of the subject. This yields
the following likelihood function:
\begin{equation}
	p(\X \cond \btheta, \boldsymbol{\pi}) = p(x_{\text{gender}} | \theta_{\text{gender}})\sum_{k=1}^K \boldsymbol{\pi}_k p(\X_{\setminus \{\text{gender}\}} \cond \btheta^{(k)}, x_{\text{gender}}).
\end{equation}
We refer to this model as ''No alive/dead strat.''.

The ARD data is an aggregate of a follow-up study which ended either on 31 December 2012 or 
on the subject's death. In this study, we are interested if an individual dies due to alcohol related 
reasons. Since the subject cannot be dead due to alcohol related reasons while still continuing to the end 
of the follow-up, we separated the model according to subjects' status by the end of the follow-up. 
This leads to the final ''Stratified'' model used in our experiments, with likelihood given as
\begin{equation}
	p(\X \cond \btheta, \boldsymbol{\pi}) = p(x_{\text{gender}} | \theta_{\text{gender}})p(x_{\text{dead}} | x_{\text{gender}}, \theta_{\text{dead}})\sum_{k=1}^K \boldsymbol{\pi}_k p(\X_{\setminus \{\text{dead, gender}\}} \cond \btheta^{(k)}, x_{\text{dead}}, x_{\text{gender}}).
\end{equation}
Here, $x_{\text{dead}}$ denotes the end of follow-up indicator and $\X_{\setminus \{\text{dead}\}}$
the features of the data excluding the end of follow-up indicator. Now we can learn two mixture
models, one for alive and the other for dead subjects for both females and males.
Since the alive subjects stay in the study until the end of the follow-up, we can model the feature
pair (''start date'', ''duration of follow-up'') using just either one of the features. In our
experiments we used the ''start date'' feature. Similarly as the ARD death can only occur in dead
subjects, we can remove this feature from the alive model.

\subsection{Similarity measures}
In the Carat experiments, we measured the performance in terms of the similarity between the covariance
matrices of the original and synthetic data. The Frobenius norm between two matrices $A$ and $B$ is given as:
\begin{equation}
	\label{eq:frob_norm}
	||A-B||_F = \left( \sum_{i=1}^n \sum_{j=1}^d (a_{ij}-b_{ij})^2  \right)^{1/2}.
\end{equation}


\subsection*{Acknowledgements}
This work was supported by the Academy of Finland; Grants 325573, 325572, 319264, 313124, 303816, 303815, 297741, 292334 
and Flagship programme: Finnish Center for Artificial Intelligence, FCAI.
We thank the Carat group for access to the Carat data (\url{http://carat.cs.helsinki.fi/})
and the CARING study group (\url{https://www.caring-diabetes.eu/}) for access to the ARD data.

\subsection*{Author contributions}
J.J., S.K., and A.H. designed research; J.J., E.L., and J.H. analyzed data; J.J., E.L., J.H., S.T., S.K. and A.H. interpreted the results; and J.J., S.K., and A.H. wrote the paper.

\subsection*{Competing interests}
The authors declare no conflict of interest.

\subsection*{Data and code availability}
Code used in our experiments is available at \url{https://github.com/DPBayes/data-sharing-examples}.

The Adult data set is available from UCI machine learning repository (\url{https://archive.ics.uci.edu/ml/datasets/adult}).

ARD and Carat data sets contain personal information and therefore are not publicly available.

Regarding the Carat data gathering process, the user is informed about the data gathering and the
research usage of the data (including app data) when installing the application in the End-user
License Agreement (EULA). The process complies with EU's General Data Protection Regulation (GDPR).
The app requires user consent for the installation. The developers have IRB (ethical board) approval
for the Carat data gathering and analysis. Anonymized subset of Carat data can be found here:
\url{https://www.cs.helsinki.fi/group/carat/data-sharing/}.

The ARD data was a collection from multiple sources: SII (permission Kela 16/522/2012),
the Finnish Cancer Registry, National Institute for Health and Welfare (THL/264/5.05.00/2012)
and Statistics Finland (TK-53-214-12). This is a register-based study with pseudonymous data and no patient
contact, thus no consents from pseudonymized patients were required according to Finnish law. The
Ethical Committee of the Faculty of Medicine, University of Helsinki, Finland (02/2012) reviewed
the protocol. Data permits were received from the Social Insurance Institute (SII) (16/522/2012),
the National Institute for Health and Welfare (THL/264/5.05.00/2012) and Statistics Finland
(TK-53-214-12). SII pseudonymized the data.

\bibliographystyle{myabbrvnat}
\bibliography{data-sharing}

\begin{thebibliography}{33}
\providecommand{\natexlab}[1]{#1}
\providecommand{\url}[1]{\texttt{#1}}
\expandafter\ifx\csname urlstyle\endcsname\relax
  \providecommand{\doi}[1]{doi: #1}\else
  \providecommand{\doi}{doi: \begingroup \urlstyle{rm}\Url}\fi

\bibitem[Abadi et~al.(2016)Abadi, Chu, Goodfellow, McMahan, Mironov, Talwar,
  and Zhang]{Abadi2016}
M.~Abadi, A.~Chu, I.~Goodfellow, H.~B. McMahan, I.~Mironov, K.~Talwar, and
  L.~Zhang.
\newblock Deep learning with differential privacy.
\newblock In \emph{Proceedings of the 2016 ACM SIGSAC Conference on Computer
  and Communications Security}, CCS '16, pages 308--318, New York, NY, USA,
  2016. ACM.

\bibitem[Abay et~al.(2018)Abay, Zhou, Kantarcioglu, Thuraisingham, and
  Sweeney]{abay2018privacy}
N.~C. Abay, Y.~Zhou, M.~Kantarcioglu, B.~Thuraisingham, and L.~Sweeney.
\newblock Privacy preserving synthetic data release using deep learning.
\newblock In \emph{Joint European Conference on Machine Learning and Knowledge
  Discovery in Databases}, pages 510--526. Springer, 2018.

\bibitem[Abowd and Vilhuber(2008)]{Abowd2008}
J.~M. Abowd and L.~Vilhuber.
\newblock How protective are synthetic data?
\newblock In \emph{Privacy in Statistical Databases}, pages 239--246, Berlin,
  Heidelberg, 2008. Springer Berlin Heidelberg.

\bibitem[{Acs} et~al.(2017){Acs}, {Melis}, {Castelluccia}, and {De
  Cristofaro}]{acs2018differentially}
G.~{Acs}, L.~{Melis}, C.~{Castelluccia}, and E.~{De Cristofaro}.
\newblock Differentially private mixture of generative neural networks.
\newblock In \emph{2017 IEEE International Conference on Data Mining (ICDM)},
  pages 715--720, Nov 2017.

\bibitem[Beimel et~al.(2010)Beimel, Kasiviswanathan, and
  Nissim]{beimel2010bounds}
A.~Beimel, S.~P. Kasiviswanathan, and K.~Nissim.
\newblock Bounds on the sample complexity for private learning and private data
  release.
\newblock In \emph{Theory of Cryptography Conference}, pages 437--454.
  Springer, 2010.

\bibitem[Blum et~al.(2008)Blum, Ligett, and Roth]{blum2008learning}
A.~Blum, K.~Ligett, and A.~Roth.
\newblock A learning theory approach to non-interactive database privacy.
\newblock In \emph{Proceedings of the Fortieth Annual ACM Symposium on Theory
  of Computing}, STOC '08, pages 609--618, New York, NY, USA, 2008. ACM.

\bibitem[But et~al.(2017)But, De~Bruin, Bazelier, Hjellvik, Andersen, Auvinen,
  Starup-Linde, Schmidt, Furu, de~Vries, et~al.]{but2017cancer}
A.~But, M.~L. De~Bruin, M.~T. Bazelier, V.~Hjellvik, M.~Andersen, A.~Auvinen,
  J.~Starup-Linde, M.~K. Schmidt, K.~Furu, F.~de~Vries, et~al.
\newblock Cancer risk among insulin users: comparing analogues with human
  insulin in the {CARING} five-country cohort study.
\newblock \emph{Diabetologia}, 60\penalty0 (9):\penalty0 1691--1703, 2017.

\bibitem[Chanyaswad et~al.(2019)Chanyaswad, Liu, and Mittal]{chanyaswad2019ron}
T.~Chanyaswad, C.~Liu, and P.~Mittal.
\newblock Ron-gauss: Enhancing utility in non-interactive private data release.
\newblock \emph{Proceedings on Privacy Enhancing Technologies}, 2019\penalty0
  (1):\penalty0 26--46, 2019.

\bibitem[Chen et~al.(2011)Chen, Mohammed, Fung, Desai, and
  Xiong]{chen2011publishing}
R.~Chen, N.~Mohammed, B.~C. Fung, B.~C. Desai, and L.~Xiong.
\newblock Publishing set-valued data via differential privacy.
\newblock \emph{Proceedings of the VLDB Endowment}, 4\penalty0 (11):\penalty0
  1087--1098, 2011.

\bibitem[Chen et~al.(2012)Chen, Acs, and Castelluccia]{chen2012differentially}
R.~Chen, G.~Acs, and C.~Castelluccia.
\newblock Differentially private sequential data publication via
  variable-length n-grams.
\newblock In \emph{Proceedings of the 2012 ACM conference on Computer and
  communications security}, pages 638--649. ACM, 2012.

\bibitem[Dua and Graff(2017)]{Dua:2019}
D.~Dua and C.~Graff.
\newblock {UCI} machine learning repository, 2017.
\newblock URL \url{http://archive.ics.uci.edu/ml}.

\bibitem[Dwork and Roth(2014)]{DworkRoth}
C.~Dwork and A.~Roth.
\newblock The algorithmic foundations of differential privacy.
\newblock \emph{Found. Trends Theor. Comput. Sci.}, 9\penalty0 (3--4):\penalty0
  211--407, August 2014.

\bibitem[Dwork et~al.(2006{\natexlab{a}})Dwork, Kenthapadi, McSherry, Mironov,
  and Naor]{dwork2006our}
C.~Dwork, K.~Kenthapadi, F.~McSherry, I.~Mironov, and M.~Naor.
\newblock Our data, ourselves: Privacy via distributed noise generation.
\newblock In \emph{Annual International Conference on the Theory and
  Applications of Cryptographic Techniques}, pages 486--503. Springer,
  2006{\natexlab{a}}.

\bibitem[Dwork et~al.(2006{\natexlab{b}})Dwork, McSherry, Nissim, and
  Smith]{dwork_et_al_2006}
C.~Dwork, F.~McSherry, K.~Nissim, and A.~Smith.
\newblock Calibrating noise to sensitivity in private data analysis.
\newblock In \emph{TCC 2006}. 2006{\natexlab{b}}.

\bibitem[Dwork et~al.(2009)Dwork, Naor, Reingold, Rothblum, and
  Vadhan]{dwork2009complexity}
C.~Dwork, M.~Naor, O.~Reingold, G.~N. Rothblum, and S.~Vadhan.
\newblock On the complexity of differentially private data release: Efficient
  algorithms and hardness results.
\newblock In \emph{Proceedings of the Forty-first Annual ACM Symposium on
  Theory of Computing}, STOC '09, pages 381--390, New York, NY, USA, 2009. ACM.

\bibitem[Dwork et~al.(2010)Dwork, Rothblum, and Vadhan]{dwork2010boosting}
C.~Dwork, G.~N. Rothblum, and S.~Vadhan.
\newblock Boosting and differential privacy.
\newblock In \emph{2010 IEEE 51st Annual Symposium on Foundations of Computer
  Science}, pages 51--60. IEEE, 2010.

\bibitem[Garfinkel and Leclerc(2020)]{garfinkel}
S.~L. Garfinkel and P.~Leclerc.
\newblock Randomness concerns when deploying differential privacy.
\newblock In \emph{Proceedings of the 19th Workshop on Privacy in the
  Electronic Society}, WPES'20, page 73–86, New York, NY, USA, 2020.
  Association for Computing Machinery.

\bibitem[Gupta et~al.(2012)Gupta, Roth, and Ullman]{gupta2012iterative}
A.~Gupta, A.~Roth, and J.~Ullman.
\newblock Iterative constructions and private data release.
\newblock In \emph{Theory of cryptography conference}, pages 339--356.
  Springer, 2012.

\bibitem[Hardt et~al.(2012)Hardt, Ligett, and McSherry]{hardt2012simple}
M.~Hardt, K.~Ligett, and F.~McSherry.
\newblock A simple and practical algorithm for differentially private data
  release.
\newblock In \emph{Advances in Neural Information Processing Systems}, pages
  2339--2347, 2012.

\bibitem[Heikkil\"{a} et~al.(2017)Heikkil\"{a}, Lagerspetz, Kaski, Shimizu,
  Tarkoma, and Honkela]{heikkila17}
M.~Heikkil\"{a}, E.~Lagerspetz, S.~Kaski, K.~Shimizu, S.~Tarkoma, and
  A.~Honkela.
\newblock Differentially private bayesian learning on distributed data.
\newblock In I.~Guyon, U.~V. Luxburg, S.~Bengio, H.~Wallach, R.~Fergus,
  S.~Vishwanathan, and R.~Garnett, editors, \emph{Advances in Neural
  Information Processing Systems}, volume~30, pages 3226--3235. Curran
  Associates, Inc., 2017.

\bibitem[J{\"a}lk{\"o} et~al.(2017)J{\"a}lk{\"o}, Dikmen, and
  Honkela]{jalko2017differentially}
J.~J{\"a}lk{\"o}, O.~Dikmen, and A.~Honkela.
\newblock Differentially private variational inference for non-conjugate
  models.
\newblock In \emph{Uncertainty in Artificial Intelligence 2017 Proceedings of
  the 33rd Conference, UAI 2017}. The Association for Uncertainty in Artificial
  Intelligence, 2017.

\bibitem[Karwa and Vadhan(2018)]{karwa2018finite}
V.~Karwa and S.~Vadhan.
\newblock Finite sample differentially private confidence intervals.
\newblock In \emph{9th Innovations in Theoretical Computer Science Conference
  (ITCS 2018)}, volume~94, page~44. Schloss Dagstuhl--Leibniz-Zentrum fuer
  Informatik, 2018.

\bibitem[Leoni(2012)]{leoni2012non}
D.~Leoni.
\newblock Non-interactive differential privacy: a survey.
\newblock In \emph{Proceedings of the First International Workshop on Open
  Data}, pages 40--52, 2012.

\bibitem[Liu and Talwar(2019)]{Liu2019}
J.~Liu and K.~Talwar.
\newblock Private selection from private candidates.
\newblock In \emph{Proceedings of the 51st Annual ACM SIGACT Symposium on
  Theory of Computing}, STOC 2019, pages 298--309, New York, NY, USA, 2019.
  ACM.

\bibitem[McSherry and Talwar(2007)]{mcsherry2007mechanism}
F.~McSherry and K.~Talwar.
\newblock Mechanism design via differential privacy.
\newblock In \emph{Annual IEEE Symposium on Foundations of Computer Science
  (FOCS)}. IEEE, October 2007.

\bibitem[Mironov(2012)]{Mironov2012}
I.~Mironov.
\newblock On significance of the least significant bits for differential
  privacy.
\newblock In \emph{Proceedings of the 2012 ACM Conference on Computer and
  Communications Security}, CCS '12, pages 650--661, New York, NY, USA, 2012.
  ACM.

\bibitem[Mohammed et~al.(2011)Mohammed, Chen, Fung, and Yu]{Mohammed}
N.~Mohammed, R.~Chen, B.~C. Fung, and P.~S. Yu.
\newblock Differentially private data release for data mining.
\newblock In \emph{Proceedings of the 17th ACM SIGKDD International Conference
  on Knowledge Discovery and Data Mining}, KDD '11, pages 493--501, New York,
  NY, USA, 2011. ACM.

\bibitem[Niskanen et~al.(2018)Niskanen, Partonen, Auvinen, and
  Haukka]{niskanen2018excess}
L.~Niskanen, T.~Partonen, A.~Auvinen, and J.~Haukka.
\newblock Excess mortality in {F}innish diabetic subjects due to alcohol,
  accidents and suicide: a nationwide study.
\newblock \emph{European Journal of Endocrinology}, 1\penalty0 (aop), 2018.

\bibitem[Oliner et~al.(2013)Oliner, Iyer, Stoica, Lagerspetz, and
  Tarkoma]{oliner2013carat}
A.~J. Oliner, A.~P. Iyer, I.~Stoica, E.~Lagerspetz, and S.~Tarkoma.
\newblock Carat: Collaborative energy diagnosis for mobile devices.
\newblock In \emph{Proceedings of the 11th ACM Conference on Embedded Networked
  Sensor Systems}, page~10. ACM, 2013.

\bibitem[Rubin(1993)]{Rubin1993}
D.~B. Rubin.
\newblock Discussion: statistical disclosure limitation.
\newblock \emph{Journal of Official Statistics}, 9\penalty0 (2):\penalty0
  461--468, 1993.

\bibitem[Xiao et~al.(2010)Xiao, Xiong, and Yuan]{xiao2010differentially}
Y.~Xiao, L.~Xiong, and C.~Yuan.
\newblock Differentially private data release through multidimensional
  partitioning.
\newblock In \emph{Workshop on Secure Data Management}, pages 150--168.
  Springer, 2010.

\bibitem[Xiao et~al.(2012)Xiao, Gardner, and Xiong]{xiao2012dpcube}
Y.~Xiao, J.~Gardner, and L.~Xiong.
\newblock Dpcube: Releasing differentially private data cubes for health
  information.
\newblock In \emph{2012 IEEE 28th International Conference on Data
  Engineering}, pages 1305--1308. IEEE, 2012.

\bibitem[Zhang et~al.(2014)Zhang, Cormode, Procopiuc, Srivastava, and
  Xiao]{Zhang}
J.~Zhang, G.~Cormode, C.~M. Procopiuc, D.~Srivastava, and X.~Xiao.
\newblock {PrivBayes}: Private data release via {B}ayesian networks.
\newblock In \emph{Proceedings of the 2014 ACM SIGMOD International Conference
  on Management of Data}, SIGMOD '14, pages 1423--1434, New York, NY, USA,
  2014. ACM.

\end{thebibliography}

\section{Supplementary Information}
\beginsupplement

\paragraph{Additional experiment on ARD data}

Table \ref{table:coef_table_k40} shows that using $40$ mixture components slightly improves the fit
for both male and female cases when compared against both private and non-private results with $10$
mixture components. 

\begin{table*}[!h]
	\centering
	\resizebox{\textwidth}{!}{%
	\begin{tabular}{llllllll}
\toprule
{} &   Coefficient & Number of cases &     Original coef. &     $\epsilon=2.0$ &     $\epsilon=4.0$ &  $\epsilon=\infty$ & $\epsilon=\infty, k=40$ \\
\midrule
0 &      OAD only &             254 &  $0.657 \pm 0.108$ &  $0.474 \pm 0.209$ &  $0.591 \pm 0.189$ &  $0.887 \pm 0.149$ &         $0.7 \pm 0.121$ \\
1 &   OAD+Insulin &              12 &  $0.873 \pm 0.304$ &   $0.846 \pm 0.44$ &  $1.074 \pm 0.427$ &  $1.124 \pm 0.366$ &        $1.12 \pm 0.257$ \\
2 &  Insulin only &             117 &   $1.68 \pm 0.135$ &  $1.085 \pm 0.312$ &  $1.313 \pm 0.293$ &  $1.521 \pm 0.206$ &       $1.587 \pm 0.153$ \\
\bottomrule
\end{tabular}

	}
	\vskip6pt

	\resizebox{\textwidth}{!}{%
	\begin{tabular}{llllllll}
\toprule
{} &   Coefficient & Number of cases &     Original coef. &     $\epsilon=2.0$ &     $\epsilon=4.0$ &  $\epsilon=\infty$ & $\epsilon=\infty, k=40$ \\
\midrule
0 &      OAD only &            1052 &  $0.435 \pm 0.049$ &  $0.502 \pm 0.152$ &   $0.538 \pm 0.12$ &  $0.532 \pm 0.089$ &       $0.523 \pm 0.061$ \\
1 &   OAD+Insulin &              66 &  $0.582 \pm 0.129$ &  $0.816 \pm 0.282$ &  $0.858 \pm 0.234$ &   $0.864 \pm 0.17$ &       $0.757 \pm 0.136$ \\
2 &  Insulin only &             480 &  $1.209 \pm 0.063$ &  $1.188 \pm 0.205$ &  $1.257 \pm 0.138$ &  $1.262 \pm 0.123$ &       $1.296 \pm 0.082$ \\
\bottomrule
\end{tabular}

	}
	\caption{\textbf{ARD study}, \textbf{ABOVE} : Females, \textbf{BELOW} : Males.
	Increasing the number of mixture components improves the fit.
	}
	 \label{table:coef_table_k40}
\end{table*}
\end{document}